\documentclass[runningheads]{llncs}
\usepackage{graphicx}
\usepackage{amssymb,subcaption,adjustbox}
\usepackage{url,cite}
\usepackage{threeparttable}
\usepackage{hyperref}

\usepackage{xcolor}
\usepackage{float}
\usepackage[misc,geometry]{ifsym}

\begin{document}
\title{ICDAR 2021 Competition on Integrated Circuit Text Spotting and Aesthetic Assessment}
\titlerunning{RRC-ICText}
%
\author{Chun Chet Ng\inst{1,2} (\Letter)\and
Akmalul Khairi Bin Nazaruddin\inst{2} \and Yeong Khang Lee\inst{2} \and Xinyu Wang\inst{3} \and Yuliang Liu\inst{4} \and Chee Seng Chan\inst{1} \and \\ Lianwen Jin\inst{5} \and Yipeng Sun\inst{6} \and Lixin Fan\inst{7}}

\authorrunning{C.C. Ng et al.}
%
\institute{University of Malaya, Malaysia \\ \and
ViTrox Technologies Sdn. Bhd., Malaysia \\
\email{ictext@vitrox.com}\\
\url{https://ictext.v-one.my/} \and The University of Adelaide, Australia \and The Chinese University of Hong Kong, China \and South China University of Technology, China \and Baidu, China \and Webank, China}
\maketitle              
\begin{abstract}

With hundreds of thousands of electronic chip components are being manufactured every day, chip manufacturers have seen an increasing demand in seeking a more efficient and effective way of inspecting the quality of printed texts on chip components. The major problem that deters this area of research is the lacking of realistic text on chips datasets to act as a strong foundation. Hence, a text on chips dataset, ICText is used as the main target for the proposed Robust Reading Challenge on Integrated Circuit Text Spotting and Aesthetic Assessment (RRC-ICText) 2021 to encourage the research on this problem. Throughout the entire competition, we have received a total of 233 submissions from 10 unique teams/individuals. Details of the competition and submission results are presented in this report. 

\end{abstract}

\section{Introduction}
Recent advances in scene text researches, including related works of text detection and recognition indicate that these research fields have received a lot of attentions, and hence pushing scene text community to progress with a strong momentum \cite{long2018scene, chen2020text}. However, this momentum is not shared in the research field of text spotting on microchip components, which attracts far less attention than scene text research. Moreover, with the 5G and Internet of Things (IoT) trend flowing into almost all technical aspects of our daily life, the demands of electronic chip components are increasing exponentially \cite{deloitte}. In line of how chip manufacturers and automated testing equipment service providers carry out chip quality assurance process, we believe now is the right time to address the text detection and recognition on microchip component problem. 

As a starting point, this competition is hosted to focus on text spotting and aesthetic assessment on microchip components or {\it ICText} in short. It includes the study of methods to spot character instances on electronic components (printed circuit boards, integrated circuits, etc.), as well as to classify the multi-label aesthetic quality of each characters. Apart from the similar end goals of both scene text and {\it ICText} to achieve a high accuracy score, the proposed methods are also expected to excel in resource constrained environments. Thus, the end solution is required to be light weight and able to carry out inference in high frame per second so that it can lower the cost of chip inspection process. Such research direction is mainly adopted by microchip manufacturers or machine vision solution providers, where one of their primary goal is to ensure that the text is printed correctly and perfectly without any defects in a timely manner.

This competition involves a novel and challenging text on microchip dataset, followed by a series of difficult tasks that are designed to test capability of the character spotting models to spot the characters and classify their aesthetic classes accurately. The submitted models are also evaluated in terms of inference speed, allocated memory size and accuracy score.

\subsection{Related Works}
Deep learning methods have been widely applied in the microchip component manufacturing processes, to reduce human dependency and focus on automating the entire manufacturing pipeline. Particularly, vision inspection and automatic optical inspection system plays an important role in this pipeline. There are a number of existing works discussing the usage of machine learning based chip component defect inspection and classification methods. For examples, \cite{ijca2015, mva2015, survey2017} are making use of traditional image processing methods to detect and classify defects on printed circuit boards. While recent works such as \cite{defect2019, opcb2019, zhang2020deep} are using deep learning based object detection models to reach similar objectives. 

However, there are a limited number of related works \cite{li2014, ocr2015, woot2017, lin_using_2019} on text detection and recognition on electronic chip components as majority of the related works focusing more on the defect detection or component recognition. For example, the authors of \cite{li2014} binarise the printed circuit board images by using a local thresholding method before text recognition and the method was tested on 860 cropped image tiles with text contents from printer circuit board. Their proposed method is separated into two stages, hence the final accuracy are subjected to their respective performance and it fails to generalise to new unseen images. Besides, the scope of \cite{ocr2015} is very limited as the authors focused on recognising broken or unreadable digits only on printed circuit boards. The involved dataset consists of only digits, 0 to 9 with a set of 50 images each, resulting in a total of 500 character images. \cite{woot2017} integrated OCR library into their printed circuit board reverse engineering pipeline to retrieve the printed part number on the housing package of an integrated circuit board. While in \cite{lin_using_2019}, they relied on convolution neural network to classify blurry or fractured characters on  1500 component images with roughly 20 characters and symbols. The aforementioned related works are able to achieve good performance in each of their own scenarios and datasets, but the results are not sufficiently convincing as these models are not tested on benchmark datasets within a controlled test environment settings.

First and foremost, there is not a large scale public dataset with thousands of chip components images available at the time of writing. This leads to the lacking of a fair and transparent platform for all the existing methods to compare and compete with each other, where researchers cannot verify that which models are superior than others. As shown by related works that aesthetic or cosmetic defects can be found commonly on electronic chip components and printed circuit boards, it is obvious that such defects occurs frequently on printed characters. Unfortunately, this problem has not derived interest as much as chip components level counterpart, and we believe that now is the right time to do so.

Additionally, \cite{long2018scene} showed that recent scene text researches are paying more attentions on the inference speed of their models because a faster model correlates to lower cost and high efficiency in production environment. Before us, not a single one of the existing Robust Reading Competitions (RRC) \cite{karatzas2013icdar, karatzas2015icdar, nayef2017icdar2017} have held similar challenges, related to deployability of deep learning models. Based on the submissions to the previous ICDAR RRC 2019 challenges with large scale datasets, especially ArT \cite{art} and LSVT \cite{lsvt}, we found out that most models are very deep and complex. They are also exposed to the weaknesses of low FPS and resource intensive models due to the nature of most submissions are coming from ensemble of multiple models. Such models are also likely to require a high amount of GPU memory during inference which make it more difficult to deploy them to end/embedded devices with low computing power.

\subsection{Motivation and Relevance to the ICDAR Community}
Text instances are usually printed on electronic components as either serial number or batch number to represent significant information. Theoretically, the printed characters should be in a simpler and cleaner format than text instances in the wild, with the assumptions that the printings are done perfectly or the images are captured under well-controlled environment. Unfortunately, this is not always the case. In production line, the quality of the printed characters is subject to various external factors, such as the changes of component material, character engraving methods, etc. Furthermore, the lighting conditions are a non trivial factor when the components are being inspected. If it is too dim then the captured images will have low contrast, while high brightness will lead to a reflection on the electronic components and yield over exposed images. Moreover, the text instances are variegated in terms of orientations, styles, and sizes across the components as different manufacturers have different requirements for character printings. Coupled with the inconsistency of printed characters with the condition of captured images, they are slightly different from what current scene text images can offer and pose problems for current text spotting models.

Convolutional neural network is used by \cite{lin_using_2019} to verify printed characters on the integrated circuit components; however, the proposed method still relies heavily on handcrafted operation and logic, which further strengthens our arguments that the aforementioned problems are valid and worth further exploration. In addition, the lack of a high quality and well annotated dataset to serve as a benchmark for stimulating works is one of the major hurdle faced by researchers in this sub-field of text spotting. Secondly, well designed tasks are also needed to fairly evaluate the proposed models in terms of the applicability and feasibility to deploy the methods for real life application, where we aim to bridge the gap between research and industrial requirements. With the proposed tasks that are focusing on the inference speed (frame per second) and also occupied GPU memory, we hope to inspire researchers to develop a light and memory efficient model. As such model is required to fit in low computing devices that are usually used in the automated optic inspection systems.

We believe that this new research direction is beneficial to the general scene text community. Through the introduction of this novel dataset and also challenges based on an industrial perspective, we aim to drive research works that are feasible to output tangible production values. With the release of this dataset, we hope to drive a new research direction towards text spotting on industrial products in the future. The proposed \textit{ICText} dataset can be served as a benchmark to evaluate existing text spotting methods. Challenges introduced in this proposed competition focus on the study of the effectiveness and efficiency of text spotting methods are also the first among previous RRC \cite{karatzas2013icdar, karatzas2015icdar, nayef2017icdar2017}, where we believe such practical requirements are crucial to drive the development of lightweight and efficient deep learning text spotting models.

\section{Competition Description}
\subsection{\textit{ICText} Dataset}
\label{dataset_info}
The novel dataset proposed for this competition, \textit{ICText} is collected by ViTrox Corporation Berhad\footnote{\url{https://vitrox.com/}} or ViTrox in short. The company is a machine vision solution provider with the focus on development and manufacturing of automated vision inspection solutions. The images are collected and characters are annotated according to the following standards:
\begin{itemize}
    \item Images in this dataset contain multi-oriented, slanted, horizontal and vertical text instances. There are also combinations of multiple text orientations.
    \item Different electronic components are involved during image collection under different lighting conditions so that \textit{ICText} represents the problems faced by the industry realistically.
    \item All text instances are annotated with tight character level bounding boxes, transcriptions and also aesthetic classes.
    \item  Sensitive information is censored wherever appropriate, such as proprietary chip designs and company information.
\end{itemize}

\subsubsection{Type/Source of Images.}
\begin{figure}[!ht]
	\centering
	\begin{subfigure}[b]{0.3\textwidth}
		\includegraphics[width=\textwidth]{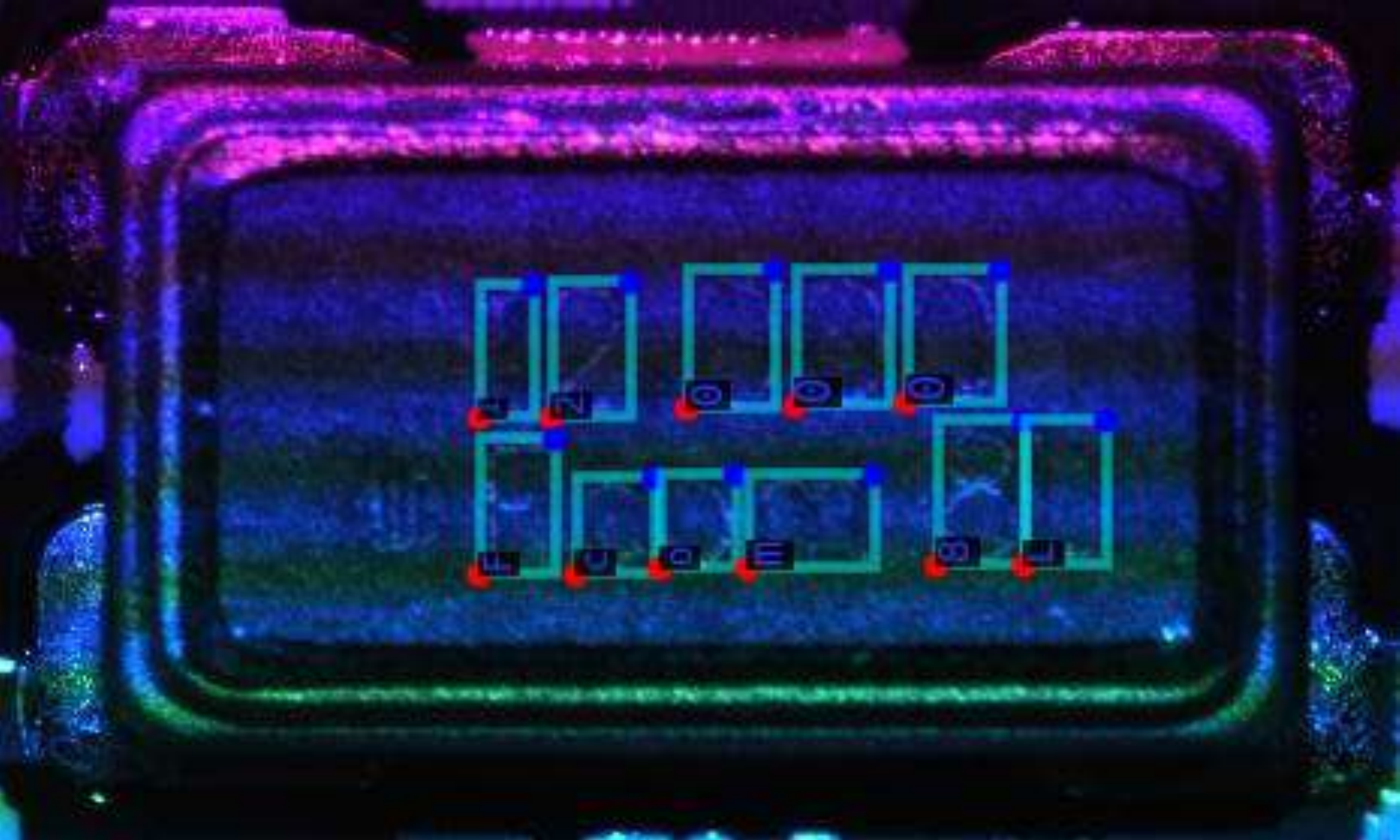}
		\caption{}
		\label{fig:sample1}
	\end{subfigure}
	\begin{subfigure}[b]{0.3\textwidth}
		\includegraphics[width=\textwidth]{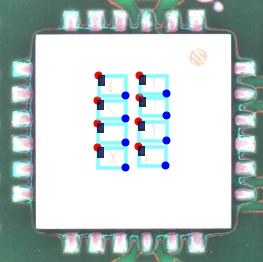}
		\caption{}
		\label{fig:sample2}
	\end{subfigure}
	\begin{subfigure}[b]{0.3\textwidth}
		\includegraphics[width=\textwidth]{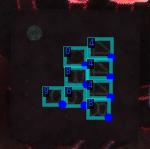}
		\caption{}
		\label{fig:sample3}
	\end{subfigure}
	\begin{subfigure}[b]{0.3\textwidth}
		\includegraphics[width=\textwidth]{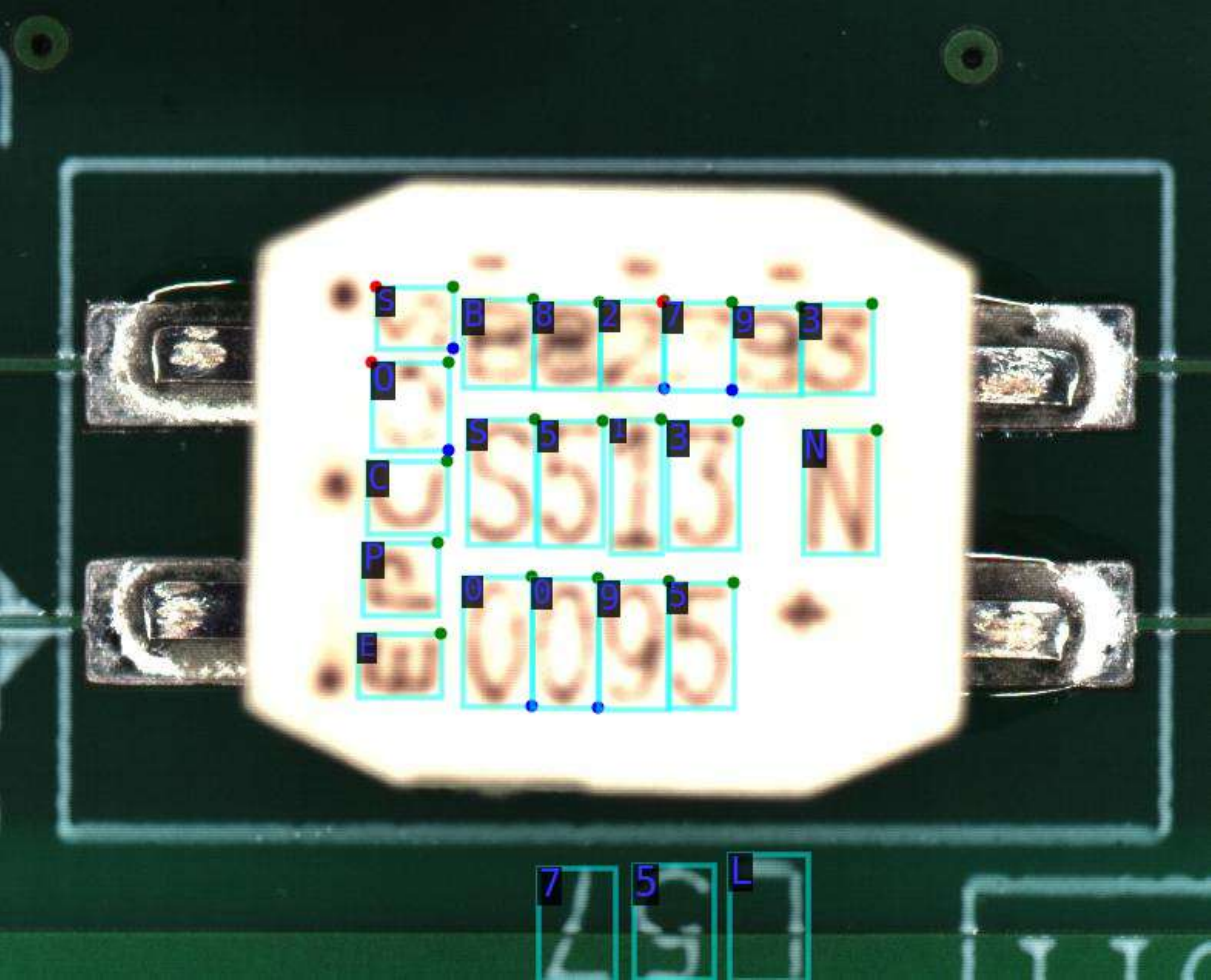}
		\caption{}
		\label{fig:sample4}
	\end{subfigure}
	\begin{subfigure}[b]{0.3\textwidth}
		\includegraphics[width=\textwidth]{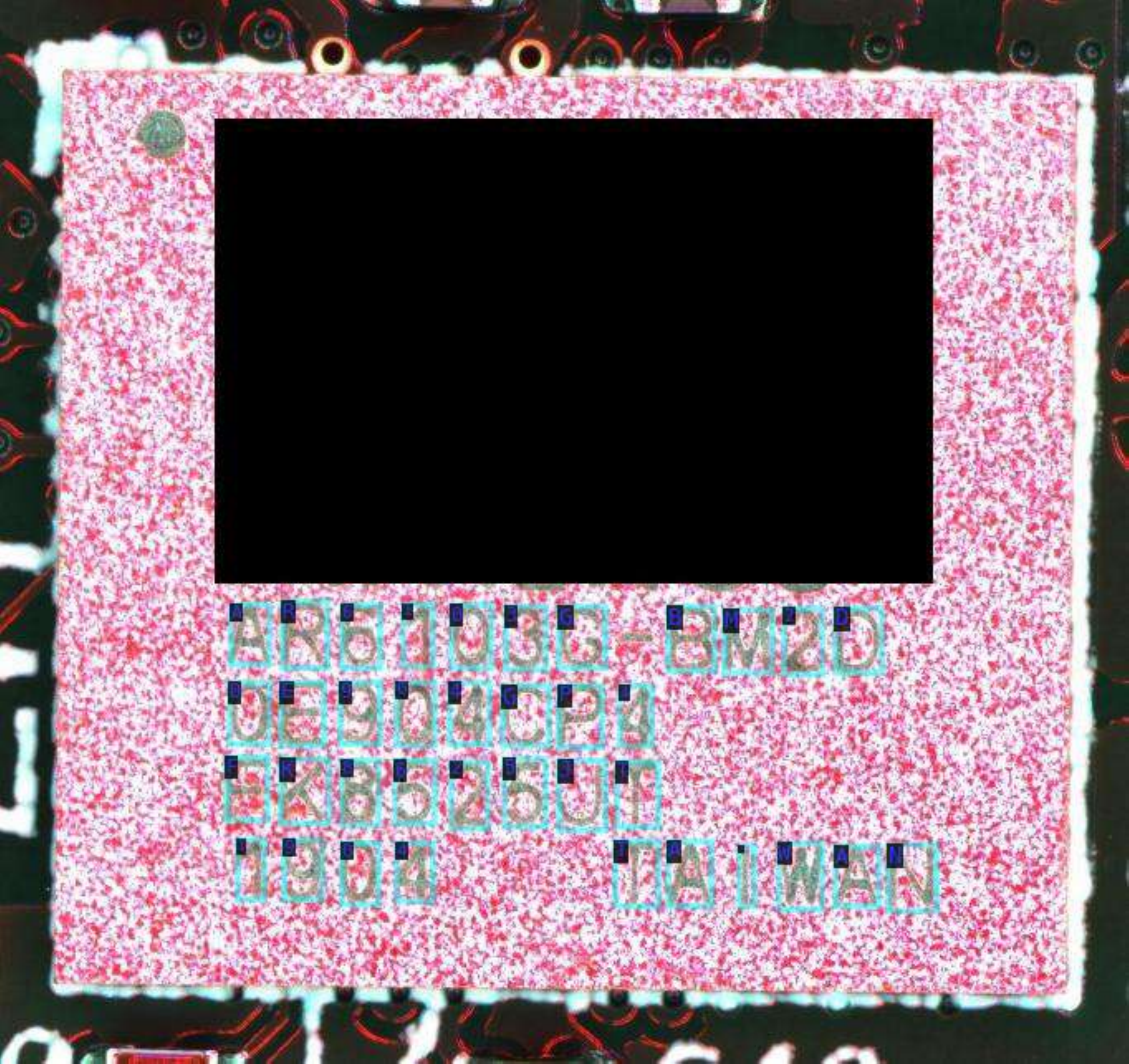}
		\caption{}
		\label{fig:sample5}
	\end{subfigure}
	\begin{subfigure}[b]{0.3\textwidth}
		\includegraphics[width=\textwidth]{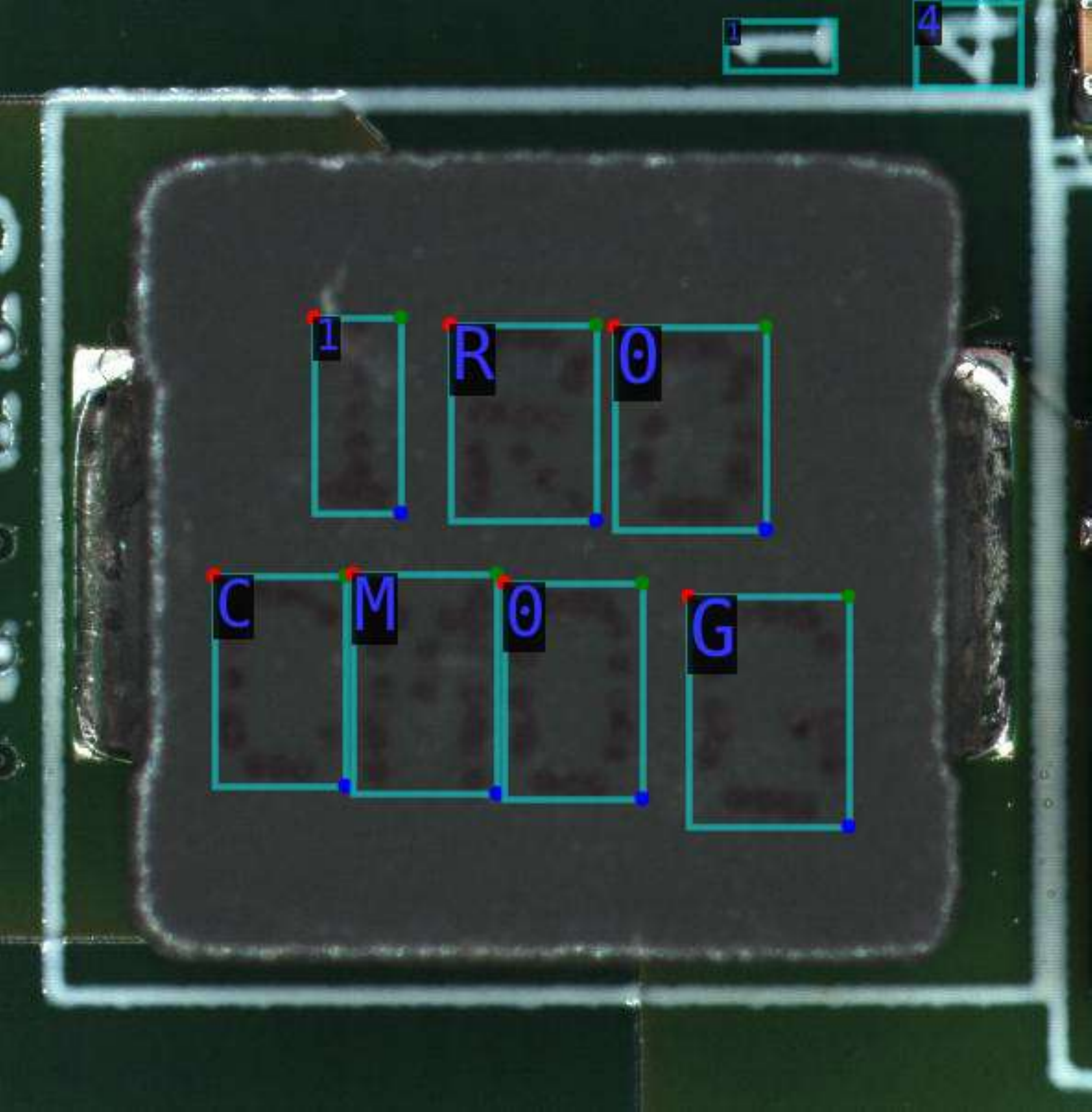}
		\caption{}
		\label{fig:sample6}
	\end{subfigure}
	\caption{Images sampled from the entire dataset. We show that they contain characters with combination of multiple font style, sizes orientations and aesthetic classes. Red dot stands for characters with low contrast compared to background; green dot represents blurry characters; blue dot highlights broken characters.}
	\label{fig:example_images}
\end{figure}

Images in \textit{ICText} dataset were collected by ViTrox's automated optical inspection machine, where the microchip components are coming from different clients. In general, images are captured from top angle with a microchip component with characters. Different automated optical inspection systems have variable configurations, leading to cases where the microchip component images are captured under different lighting conditions. We can also notice that certain images are under-exposed and over-exposed due to these settings. Sample images can be found in Figure \ref{fig:sample1}, \ref{fig:sample2} and \ref{fig:sample3}, where we can observe a low contrast between the printed characters and the image background, making it very difficult to spot the characters on the images. Apart from the usual vision-related challenges (illumination, blurriness, perspective distortion, etc.), \textit{ICText} stands out in challenging scene text understanding models with the combination of different text orientations in one single image and never seen before background (different microchip components). As illustrated in Figure \ref{fig:sample4}, \ref{fig:sample5} and \ref{fig:sample6}, the multi-oriented characters are incomplete/broken with separated body/branches and also inseparable with the background. The variation of chip component backgrounds indirectly affect the quality of printed characters and leading to multiple font styles on the images. These uncontrollable defects are subjected to the character engraving methods, chip component's base materials or printing defects during manufacturing. 

\subsubsection{\textit{ICText} versus Scene Text Dataset.}\textit{ICText} includes different type of images as compared to existing scene text datasets, where the major difference lies in the image background. Characters in \textit{ICText} are printed on microchip components, whereas text instances in scene text dataset have much more complex and vibrant colour backgrounds. Because the images in \textit{ICText} are captured under different camera settings, there exist cases where some images have extreme illuminations conditions (low-contrast versus over-exposure) and image quality (blurry versus well-focused). We believe that such characteristic are challenging to text spotting models, especially in the cases where the text and background colour has low contrast. Besides, printed characters in our chip component images involve a huge variety of font styles and sizes due to the fact that different manufacturers have different requirements in character printing. Hence the challenging aspect of font variations in scene text images are also well observed in \textit{ICText}. The diversity of images are ensured by collecting images of components with different structures and text orientations. To prevent the same set of components with similar structures or characters exist in between training and testing set, we carry out careful filtering on the collected images.

\subsubsection{\textit{ICText} Dataset Splits.}There are a total of 20,000 images in the \textit{ICText} dataset. It is split into a training set (10,000 images), validation set (3000 images) and testing set (10,000 images). Images and annotations for training set will be released by on-demand request basis, and a Non Disclosure Agreement has to be signed prior to the release. However, the annotations for the validation set and the entire test set are kept private due to privacy and copyright protection. A comparison between \textit{ICText} dataset and other related datasets can be found in Table \ref{Dataset_table}, where we can see that \textit{ICText} is the most complete and largest text on chips dataset at the time of writing.

\addtolength{\tabcolsep}{5pt} 
\begin{table*}
\caption{A quantitative comparison between \textit{ICText} and other microchip component text datasets. All four datasets report the presence of character defects (low contrast, blurry text, broken text).} \label{Dataset_table}
\begin{center}
    \begin{tabular}{|l|c|c|c|}
    \hline
    Year & Paper & Number of Images \\
    \hline
    2014 & \cite{li2014} & 860 cropped text images.\\
    2015 & \cite{ocr2015} & 500 character images. \\
    2019 & \cite{lin_using_2019} & 1,500 component images.\\
    2021 & \textit{ICText} & 20,000 component images.\\
    \hline
    \end{tabular}
\end{center}
\end{table*}

\subsubsection{Ground truth Annotation Format.}All text instances in the images are annotated at the character level with a tight bounding box. \textit{ICText} follows the MS COCO's annotation JSON format for object detection. There is an entry for each image in the dataset and match with their respective annotation entries. Apart from the usual keys in MS COCO's JSON format, we also provide additional information at the character level, such as the illegibility, degree of rotation and aesthetic classes. The ``bbox" key represents the 4 pairs of polygon coordinates, and the ``rotation\_degree" key represents the angle the image can be rotated to get human-readable characters. The counter-clockwise rotation angle is provided to ease the training of models, so that rotation sensitive characters like 6 and 9 will not be confused. Multi labels for aesthetic classes are one-hot encoded, where each column is a class. Similar to COCO-text \cite{gomez2017icdar2017} and ICDAR2015 \cite{karatzas2015icdar}, annotations marked with ignore flag are do not care regions, they will not be taken into consideration during evaluation.

\subsection{Competition Tasks}
\label{task_detail}
The RRC-\textit{ICText} consists of three main tasks: text spotting, end-to-end text spotting and aesthetic assessment, and lastly inference speed, allocated memory size and accuracy score assessment. This section details them.

\subsubsection{Task 1: Text Spotting.}The main objective of this task is to detect the location of every character and then recognise it given an input image, which is similar to all the previous RRC scene text detection tasks. The input modal of this task is strictly constrained to image only, no other form of input is allowed to aid the model in the process of detecting and recognising the characters. \\
\noindent \textbf{Input:} Microchip component text image. \\
\textbf{Output:} Bounding box with its corresponding class (a-z, A-Z, 0-9).

\subsubsection{Task 2: End-to-end Text Spotting and Aesthetic Assessment.}The main objective of this task is to detect and recognize every character and also their respective multi-label aesthetic classes in the provided image. \\
\noindent\textbf{Input:} Microchip component text image. \\
\textbf{Output:} Bounding box with the corresponding class/category (a-z, A-Z, 0-9) and multi-label aesthetic classes.

\subsubsection{Task 3: Inference Speed, Model Size and Score Assessment.}The main objective of this task is to evaluate the submitted models on their feasibility and applicability of deploying the methods for real life applications. In order to evaluate the submitted methods under previous tasks (Task 1 and 2), we decided to separate this task into two sub-tasks:\\
\noindent\textbf{Task 3.1} - This sub-task takes the score of \textbf{Task 1} into consideration, we intend to evaluate the speed, size and text spotting score of the submitted model.\\
\textbf{Task 3.2} - This sub-task is meant to evaluate the model on inference speed, occupied GPU memory size and multi-label classification score as stated in \textbf{Task 2}.\\

\noindent Submission to these two tasks are evaluated on the private test set through a docker image with inference code, model and trained weights. This is to study the relationship between speed, size, and score, where the end goal is to research and develop a well balanced model based on these three aspects. The submitted model is expected to achieve an optimal performance with minimal inference time and occupied GPU memory, a representation of efficiency and effectiveness.\\
\noindent \textbf{Input:} Docker image with inference code, model and trained weights. \\
\textbf{Output:} Inference speed, model size and respective Score.

\subsection{Evaluation Protocol}
\label{eval_proc}
Evaluation servers for all the tasks are prepared at Eval.ai with separate pages for Task 1 \& 2\footnote{\url{https://eval.ai/web/challenges/challenge-page/756/overview}} and Task 3\footnote{\url{https://eval.ai/web/challenges/challenge-page/757/overview}}. Do note that due to privacy concerns and prevention of malicious use of the \textit{ICText} dataset, the Task 1 \& 2 are evaluated on validation set and only Task 3 is evaluated on the entire private test set. Relevant evaluation code can be found at this GitHub repository\footnote{\url{https://github.com/vitrox-technologies/ictext_eval}}.

\subsubsection{Evaluation metric for Task 1.}We follow the evaluation protocol of the MSCOCO dataset \cite{mscoco}. As we have separated the characters to 62 classes (a-z, A-Z, 0-9), MSCOCO's evaluation protocol can give us a better insight than the commonly used Hmean score in scene text spotting. We rank the submissions based on mean Average Precision or mAP as to MSCOCO challenges.

\subsubsection{Evaluation metric for Task 2.}We first evaluate whether the predicted character matches the ground-truth at IoU 0.5 and then only we calculate precision, recall and F-2 Score value for the predicted multi-label aesthetic classes. If the predicted character is wrong, then the respective aesthetic classes will not be evaluated. A default value of [0,0,0] is given to predictions that have no aesthetic labels and we flip the aesthetic labels of both ground truth and prediction when all the values are zero, to prevent zero division error when calculating precision and recall. The submitted results are ranked by the average of multi-label F-2 Score across all images. Besides, F-2 Score is selected because recall is more important than precision when it comes to defect identification during chip inspection, as missed detection would yield high rectification costs. 



\subsubsection{Evaluation metric for Task 3.}We evaluate the submitted model in terms of inference speed (in frame per second or FPS) and also occupied GPU memory size (in megabytes or MB) together with the respective score of each sub-tasks. The evaluation metric of this task is named as Speed, Size and Score or $3S$ in short. It is defined using the following equation:

\begin{equation}\label{3S}
3S = 0.2 \times normalised\_speed + 0.2 \times (1-normalised\_size) +  0.6 \times normalised\_score
\end{equation}

We decided to give a higher priority to the respective task's score, because we believe that an optimal model should have a solid performance as a foundation, as well as able to strike a balance between speed and size. Firstly, $normalised\_speed$ of the submitted model is calculated using the following equation:
\begin{equation}
normalised\_speed = \min(\frac{actual\_FPS}{acceptable\_FPS}, 1)
\end{equation}
We execute the inference script in the submitted Docker image and record total time taken for the model to carry out inference on the entire test set, which is then used to calculate the $actual\_FPS$. For the $acceptable\_FPS$, we decided to set it to 30FPS due to the fact that it is an average FPS recorded in modern object detectors on GPU and edge computing devices like Nvidia's Jetson Nano \cite{nvidia_developer_2019} and Intel's Neural Compute Stick 2 \cite{openvino}. Furthermore, we calculate $normalised\_size$ of the submitted model using the following equation:
\begin{equation}
normalised\_size = \min(\frac{allocated\_memory\_size}{acceptable\_memory\_size}, 1)
\end{equation}
The $allocated\_memory\_size$ is defined based on the maximum GPU memory usage recorded during model inference. Then, we set the $acceptable\_memory\_size$ to 4000MB or 4GB because it is an average memory size available to mobile and edge computing devices. The $normalised\_score$ is referred to mean Average Precision for Task 3.1 and multi-label F-2 Score for Task 3.2, both scores have to be normalized to the range of 0 to 1. We then rank the participants based on the final $3S$ score from the highest to lowest. All submitted docker images to Task 3 are evaluated on the Amazon EC2 p2.xlarge instance.

\section{Results and Discussion}
At the end of the competition, we have received a total of 233 submissions from 10 unique teams/individuals. In Task 1, there are 161 submissions from all 10 teams, whereas we received 50 submissions from 3 teams in Task 2. While in Task 3.1, there are 20 submissions from 2 teams and only 1 submissions for Task 3.2. Due to a high number of repetitive submissions from the participants, we only consider the one with the best result out of all submissions from the same team in this report. The final rankings are detailed in the Table \ref{ranking_table}. 

\subsection{Result Analysis}
Based on the results, we are able to analyse the average performance of participants on \textit{ICText}, as this would help us to understand the overall upper threshold scores of all tasks. Firstly, we observe that the participants achieve an average mAP score of 0.41 in Task 1. They have an Average Precision (AP) score of 0.551 and 0.505 for Intersection over Union (IoU) at 0.5 and 0.75. With the highest mAP score of 0.60, we believe that this shows how challenging \textit{ICText} dataset is given that the participants' results are only slightly higher than the average mAP score. Besides, the average multi-label precision and recall score of participants in Task 2 is 0.42 and 0.55, with the average multi-label F-2 score reaching 0.49. As the number of submissions for Task 3.1 and 3.2 are limited, we report the average of FPS and GPU memory usage (in Megabytes) for both tasks, which are 21.14 and 3860.8 MB. 

\begin{figure*}
	\begin{subfigure}[b]{0.3\textwidth}
	\setlength\lineskip{10pt}
	    \centering
		\includegraphics[width=0.9\textwidth]{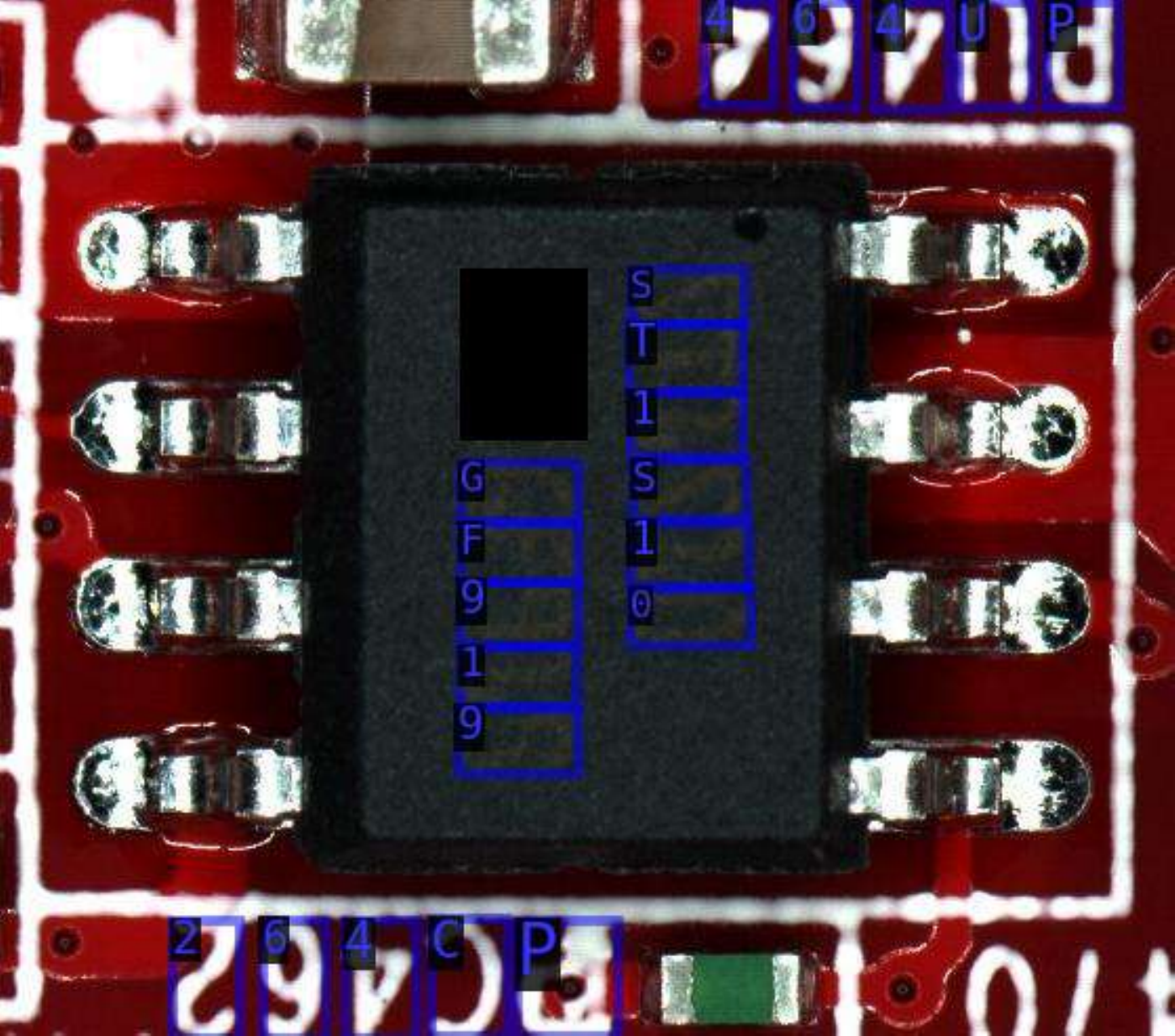}
		\includegraphics[width=0.9\textwidth]{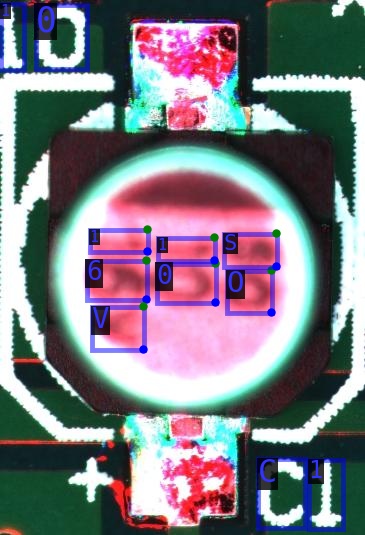}
    	\caption{Ground truth images.}
		\label{fig:infer_gt}
	\end{subfigure}
	\hfill
	\begin{subfigure}[b]{0.3\textwidth}
	\setlength\lineskip{10pt}
	    \centering
		\includegraphics[width=0.9\textwidth]{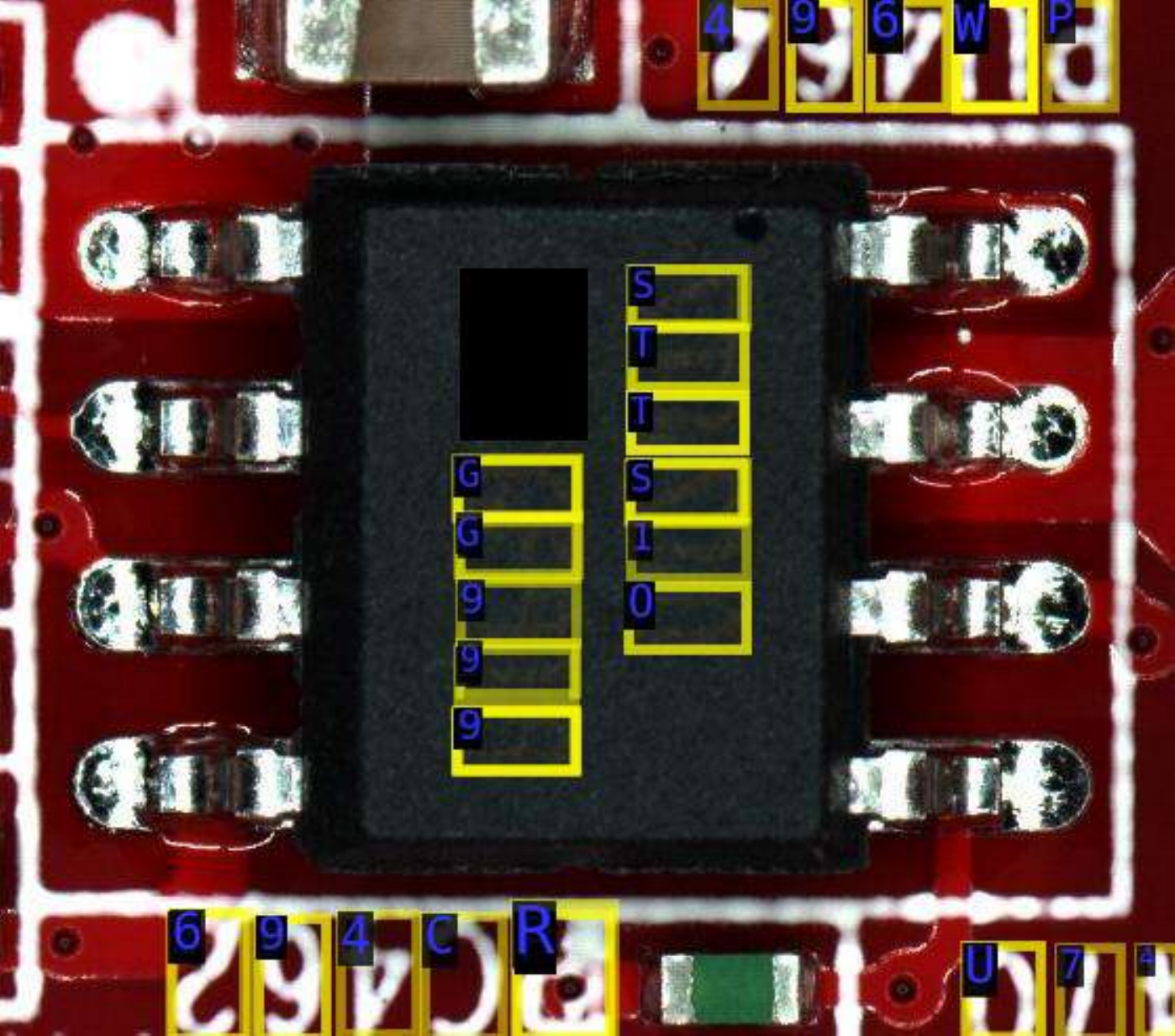}
		\includegraphics[width=0.9\textwidth]{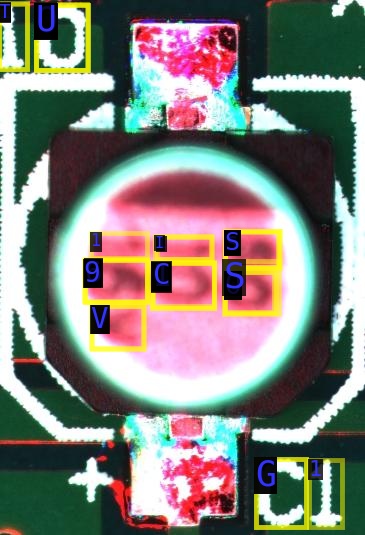}
		\caption{\textbf{gocr} team.}
		\label{fig:infer_gocr}
	\end{subfigure}
	\hfill
	\begin{subfigure}[b]{0.3\textwidth}
	\setlength\lineskip{10pt}
	    \centering
		\includegraphics[width=0.9\textwidth]{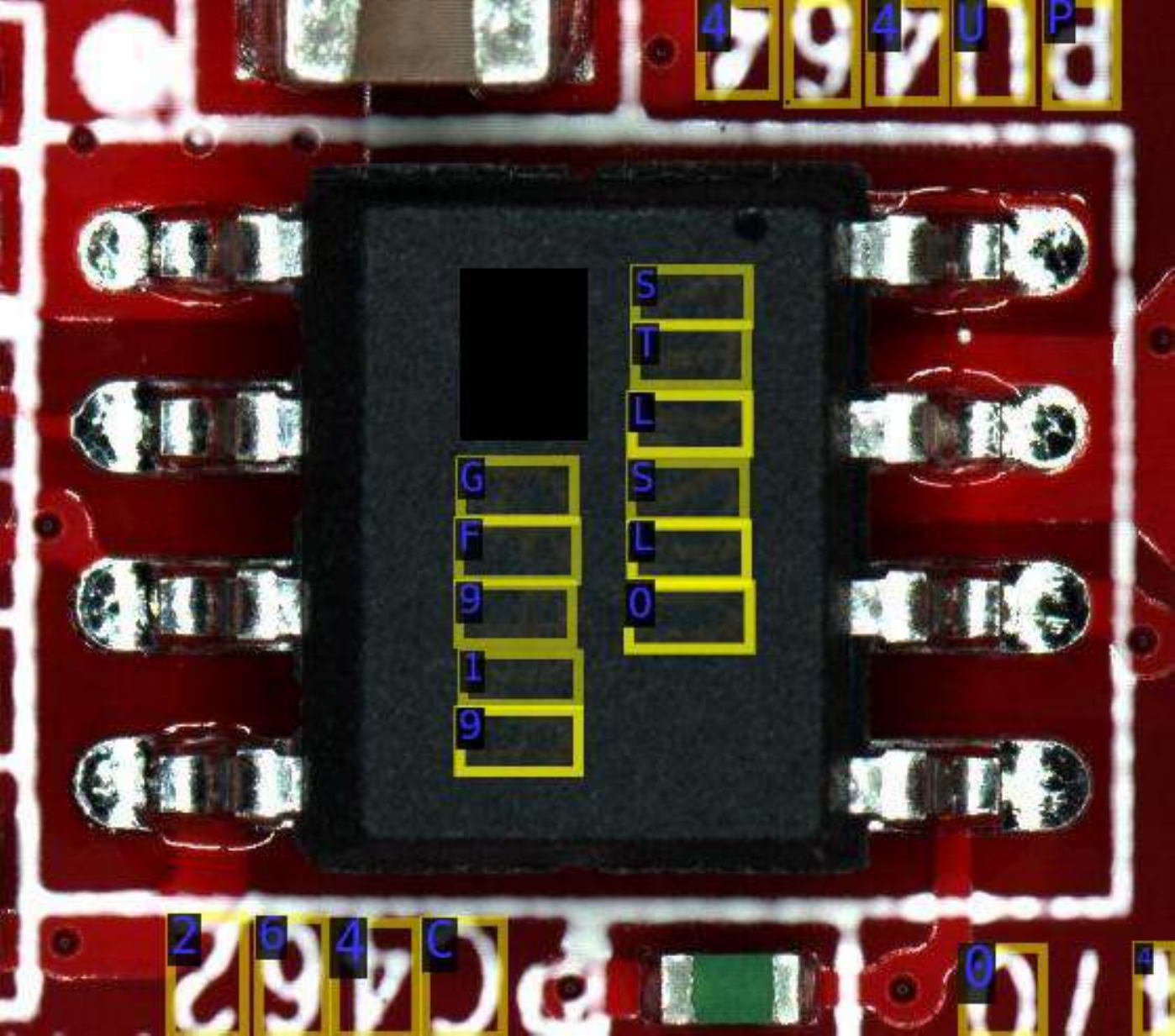}
		\includegraphics[width=0.9\textwidth]{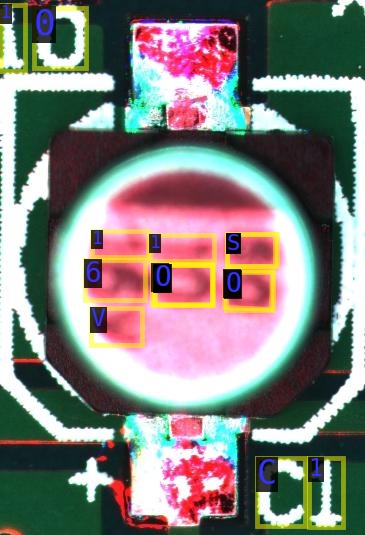}
		\caption{\textbf{SMore-OCR} team.}
		\label{fig:infer_smore}
	\end{subfigure}
	\caption{A comparison between ground truth and predictions by \textbf{gocr} and \textbf{SMore-OCR} team for Task 3.1 on the test set. Ground truth bounding boxes are in blue colour, while predicted bounding boxes are plotted in yellow.}
	\label{fig:infer_samples}
\end{figure*}

In Task 2, it is worth noting that the score of \textbf{shushiramen} team is higher than \textbf{hello\_world} team, although the latter team achieves better score in Task 1, with \textbf{shushiramen} team scoring 0.31 for mAP and \textbf{hello\_world} team has the mAP score of 0.46. This shows that the multi-label classification capability of the models are equally important as the detectors, although the predicted multi-label aesthetic classes are evaluated sequentially when the character is localised and recognized accurately. Given marginally higher mAP of \textbf{hello\_world} team as compared to the \textbf{shushiramen} team, the multi-label F-2 Score of \textbf{hello\_world} team is being dragged down by poor classification module, which will be discussed in Section \ref{participants_method}. Moreover, we also notice how the evaluation metric of Task 3.1 plays its parts on emphasizing the deployability of models on the results of \textbf{SMore-OCR} and \textbf{gocr}, where their difference in mAP is 0.06, but their 3S score is different by a margin of 50\%. This illustrates that the model of \textbf{gocr} team is more suitable to be deployed to end/edge devices as it requires less computing resources and has high inference speed. Some qualitative examples are shown in Figure \ref{fig:infer_samples}, where we observe that the models of both \textbf{gocr} (Figure \ref{fig:infer_gocr}) and \textbf{SMore-OCR} (Figure \ref{fig:infer_smore}) are unable to spot multi-oriented characters accurately with the majority of predictions having wrong character classes, and their models fail drastically when the printed characters have aesthetic defects.

\begin{table*}
\centering
\caption{Final ranking table for all the tasks in RRC-\textit{ICText} 2021.} \label{ranking_table}
\begin{center}
    \resizebox{\textwidth}{!}{
    \begin{tabular}{|l|c|c|c|c|c|}
    \hline
    \textbf{Rank} & \textbf{Team (Affiliation)} & \multicolumn{3}{c|}{\textbf{Scores}} \\
    \hline
    \multicolumn{2}{|c|}{\textbf{Task 1 - Validation Set}} & \textbf{mAP} & \textbf{AP IOU@0.5} & \textbf{AP IOU@0.75} \\
    \hline
    1 & gocr (Hikvision) & 0.60 & 0.78	& 0.73 \\
    2 & SMore-OCR (SmartMore) & 0.53 & 0.75 & 0.65 \\
    3 & VCGroup (Ping An) & 0.53 & 0.72 & 0.65 \\
    4 & war (SCUT) & 0.48 & 0.60 & 0.58 \\
    5 & hello\_world (Uni. of Malaya) & 0.46 & 0.59 & 0.56 \\
    6 & maodou (HUST) & 0.44 & 0.58 & 0.54 \\
    7 & WindBlow (Wuhan Uni.) & 0.43 & 0.57 & 0.53 \\
    8 & shushiramen (Intel) & 0.31 & 0.45 & 0.39 \\
    9 & XW (-) & 0.21 & 0.31 & 0.27 \\
    10 & Solo (-) & 0.12 & 0.16 & 0.15 \\
    \hline
    \multicolumn{2}{|c|}{\textbf{Task 2 - Validation Set}} & \textbf{Precision} & \textbf{Recall} & \textbf{F-2 Score} \\
    \hline
    1 & gocr (Hikvision) & 0.75 & 0.70 & 0.70 \\
    2 & shushiramen (Intel) & 0.30 & 0.65 & 0.51 \\
    3 & hello\_world (Uni. of Malaya) & 0.22 & 0.29 & 0.26 \\
    \hline
    \multicolumn{2}{|c|}{\textbf{Task 3.1 - Test Set}} & \textbf{mAP} & \textbf{FPS;GPU Mem.(MB)} & \textbf{3S} \\
    \hline
    1 & gocr (Hikvision) & 0.59 & 31.04;305.88 & 0.74 \\
    2 & SMore-OCR (SmartMore) & 0.53 & 2.69;10970.75 & 0.33 \\
    \hline
    \multicolumn{2}{|c|}{\textbf{Task 3.2 - Test Set}} & \textbf{F-2 Score} & \textbf{FPS;GPU Mem.(MB)} & \textbf{3S} \\
    \hline
    1 & gocr (Hikvision) & 0.79 & 29.68;305.88 & 0.85 \\
    \hline
    \end{tabular}}
\end{center}
\end{table*}

\subsection{Participants' Methods}
\label{participants_method}

\subsubsection{gocr (Hikvision).}Being the winning team for all three tasks, \textbf{gocr} team presents an interesting solution. Extra 84,400 images have been generated synthetically using SynthText \cite{Gupta16} and text style transfer method introduced in \cite{edit_text} to tackle the lower case character class imbalance problem in our training set. YoloV5s \cite{yolov5} is used as the main model for character spotting and an additional classification branch of 3 aesthetic classes are added to it, so that the model is catered for all tasks. In short, there are a few tips and tricks leading to the superior performance of this model, including the use of synthetic images for pre-training and fine tuning, generation of pseudo labels based on validation images for semi supervised learning and the employment of teacher-student learning strategy for knowledge distillation. The team has also taken additional measures to deploy their model using TensorRT so that it can achieve faster speed and utilize less GPU resources, which is well aligned with our objectives in Task 3. We recommend readers to check out their main paper \cite{gocr} for more information.

\subsubsection{SMore-OCR (SmartMore).}Mask-RCNN \cite{maskrcnn} based character spotting model has been proposed by this team. Before the model's training commences, the team generated 30,000 synthetic images based on our training set. They first mask out the training set images by using the provided bounding boxes, and then the EdgeConnect method \cite{Nazeri_2019_ICCV} is used to fill up the blanks. After that, they follow SynthText's methodology \cite{Gupta16} to write characters back to the synthetic images. Once the process is completed, they pre-trained a Mask R-CNN model on the synthetically generated data, before fine tuning on our training set. The Mask R-CNN is initialized with default hyper-parameters and trained for 6 epochs with the batch size of 32. In order to recognize multi-oriented characters based on the output of Mask-RCNN, two ImageNet pre-trained ResNet-50 are fine-tuned on the synthetic data and our training set as the character orientation and classification models. Both models use mini-batch stochastic gradient decent (SGD) with momentum of 0.9 and weight decay of $5 \times 10^-4$. The learning rate is dynamically adjusted during the training process with the following equation of $lr = lr_0 \div (1 + \alpha{p})^\tau$, where $lr_0 = 10^{-3}$, $\alpha = 10$ and $\tau = 0.75$, with $p$ being linearly adjusted from 0 to 1. However, the team claims that the addition of these two models do not improve the end results and subsequently drops it out of the entire pipeline, hence the final results are solely based on Mask-RCNN.

\subsubsection{hello\_world (Uni. of Malaya).}The MS-COCO pre-trained YoloV5m \cite{yolov5} is selected as the main model, which is fine-tuned on our dataset for 120 epochs with the settings of 576 as the image size and using the batch size 40. Besides, the default learning rate is used and the training data is augmented using mosaic image augmentation. However, the character rotation angle information is not used during training. In order to improve the character spotting performance, the team generates synthetic binary images of random characters as extra training data. Furthermore, a multi-label aesthetic model is separately trained using ImageNet pre-trained mobilenet\_v2 on the cropped 64x64 character bounding box images for 4 epochs with binary cross entropy loss. During inference, the fine-tuned YoloV5m first detects, recognizes the characters, crops them out and passes to the aesthetic model for multi-label classifications.


\subsection{Error Analysis}
Following the method introduced in \cite{bolya2020tide}, TIDE is selected as the main error analysis tool because it provides insightful explanations on the weaknesses of participants' models and able to pinpoint the specific errors that are commonly made on our \textit{ICText} dataset. Hence, TIDE is used to analyse all submissions to Task 1 and Task 3.1. Results are shown in Table \ref{tide_table}, where the values in the table are mAP scores in the range of 0 to 100 if that particular error is fixed. We determine that the rate of occurrence for a specific error is high when the mAP increment is high. In this work, we focus on the main error types, as the special error types (FalsePos and FalseNeg) might have overlapping cases with them.

We observe that the most common error made by all models is the Classification error (Cls). We find it interesting that the mAP increment of lower ranking models are a few times higher than the top 3 models, this shows that they are performing badly to classify the detected characters. On the other hand, out of the 484285 predictions by the \textbf{gocr} team's Yolov5s model, 41\% of them or 200694 predictions are categorised under Classification error. With this high number of predictions, we can also see that the model contributes the most to False Positive error (FalsePos). This scenario happens when the model might confuse between similar characters such as 6 or 9, and even in between case-insensitive characters like v and V. While the Mask R-CNN used by \textbf{SMore-OCR} team produce 62054 bounding boxes, which is 7.8 times less than what \textbf{gocr} team outputted. And only 18148 of them are categorised under Classification error, constituting 29\% of the total predictions. This huge difference on number of predictions can be explained by the difference in adopted model architecture by these two teams, where Yolov5 is a one-stage object detection model that produces anchor boxes at more locations in the image compared to the two stage Mask R-CNN model, where the number of anchor boxes is limited. Furthermore, the second most frequent error made by the participants' models is Missing error, where the models fail to detect certain ground truth characters. The \textbf{Solo} team's model suffers the most under this error, and it can only matched 8059 out of 30400 ground-truth bounding boxes, which is only 26.5\%. Coupling with the high False Negative error, this helps to explain why the model performs badly on our dataset. Based on the results reported on both Task 1 and Task 3.1 in Table \ref{tide_table}, we can conclude that all models are facing difficulties to accurately recognize multi-oriented and defective characters, which are the two main aspects of \textit{ICText} dataset.

\begin{table*}
\centering
\caption{TIDE results of all submissions in Task 1 and Task 3.1.} \label{tide_table}
\begin{center}
    \resizebox{\textwidth}{!}{
    \begin{tabular}{|l|c|c|c|c|c|c|c|c|}
    \hline
    \textbf{Team} & \textbf{Cls} & \textbf{Loc} & \textbf{Both} & \textbf{Dupe} & \textbf{Bkg} & \textbf{Miss} & \textbf{FalsePos} & \textbf{FalseNeg} \\
    \hline
    \multicolumn{9}{|c|}{\textbf{Task 1 - Validation Set}}\\
    \hline
    gocr & 7.27 & 0.2 & 0.02 & 0 & 5.44 & 1.06 & 14.86 & 3.65 \\
    SMore-OCR & 11.14 & 0.67 & 0.14 & 0 & 6.12 & 1.94 & 10.37 & 9.05\\
    VCGroup & 6.01 & 0.38 & 0.09 & 0 & 5.49 & 4.97 & 12.99 & 8.77\\
    war & 28.07 & 0.14 & 0.03 & 0 & 2.09 & 2.56 & 6.99 & 17.72\\
    hello\_world & 32.08 & 0.18 & 0.02 & 0 & 2.52 & 0.85 & 4.81 & 14.32\\
    maodou & 33.44 & 0.09 & 0.02 & 0 & 1.72 & 1.85 & 3.36 & 29.26\\
    WindBlow & 31.77 & 0.07 & 0.01 & 0 & 1.7 & 3.1 & 3.39 & 28.97\\
    shushiramen & 42.55 & 0.25 & 0.23 & 0.01 & 1.41 & 2.44 & 10.02 & 30\\
    XW & 40.53 & 0.22 & 0.37 & 0 & 0.93 & 6.58 & 4.87 & 29.02\\
    Solo & 11.97 & 0.03 & 0.01 & 0 & 0.28 & 38.74 & 0.94 & 77.94\\
    \hline
    \multicolumn{9}{|c|}{\textbf{Task 3.1 - Test Set}}\\
    \hline
    gocr & 10.85 & 0.21 & 0.03 & 0 & 0.7 & 5.67 & 7.58 & 9.87\\
    SMore-OCR & 11.14 & 0.17 & 0.12 & 0 & 1.8 & 5.92 & 6.56 & 13.45\\
    \hline
    \end{tabular}}
\end{center}
\end{table*}

\section{Conclusion}
In summary, the RRC-ICText has introduced a new perspective to the scene text community with the focus on industrial OCR problems and a never seen before large scale text on integrated circuits dataset. The competition includes challenges that represent problems faced by component chip manufacturers, involving the need to recognize characters and classify multi-label aesthetic defects accurately in a timely manner under limited computing resources. The proposed \textit{ICText} dataset, tasks, and evaluation metrics present an end-to-end platform for text spotting researchers to evaluate their methods for performance and production deployability. Results by the participants show that there is still a huge room for improvement on the dataset, albeit creative ideas have been applied to tackle the posed problems. Lastly, the RRC-ICText is a good starting point to inspire researches on scene text models that are not only emphasizing high accuracy but at the same time have low computing requirements.
%
%
%
%
%
\bibliographystyle{splncs04}
\bibliography{egbib}
\end{document}